\documentclass[conference]{IEEEtran}
\IEEEoverridecommandlockouts
\usepackage{cite}
\usepackage{amsmath,amssymb,amsfonts}
\usepackage{algorithmic}
\usepackage{graphicx}
\usepackage{textcomp}
\usepackage{xcolor}
\usepackage{hyperref}
\usepackage{balance}
\usepackage{caption}
\usepackage{float}
\usepackage{placeins} 
\usepackage{balance}

\def\BibTeX{{\rm B\kern-.05em{\sc i\kern-.025em b}\kern-.08em
    T\kern-.1667em\lower.7ex\hbox{E}\kern-.125emX}}
\begin{document}

\title{AMSnet 2.0: A Large AMS Database \\with AI Segmentation for Net Detection

\thanks{
\noindent This work was partially supported by ``Science and Technology Innovation in Yongjiang 2035" (2024Z283) and by research support from BTD Inc. \\ \indent \IEEEauthorrefmark{1}Equal contribution \\ \indent \IEEEauthorrefmark{7}Corresponding authors}
}

\author{
\IEEEauthorblockN{
    Yichen Shi\IEEEauthorrefmark{1}\IEEEauthorrefmark{4}, 
    Zhuofu Tao\IEEEauthorrefmark{1}\IEEEauthorrefmark{2},
    Yuhao Gao\IEEEauthorrefmark{4}, 
    Li Huang\IEEEauthorrefmark{4},
    Hongyang Wang\IEEEauthorrefmark{4}}
\IEEEauthorblockN{
    Zhiping Yu\IEEEauthorrefmark{5}, 
    Ting-Jung Lin\IEEEauthorrefmark{4}\IEEEauthorrefmark{6},
    Lei He \IEEEauthorrefmark{4}\IEEEauthorrefmark{6}}
\IEEEauthorblockA{\IEEEauthorrefmark{4}Ningbo Institute of Digital Twin, Eastern Institute of Technology, Ninngbo, China}
\IEEEauthorblockA{\IEEEauthorrefmark{2}University of California, Los Angeles, USA}
\IEEEauthorblockA{\IEEEauthorrefmark{5}Tsinghua University, Beijing, China}
\IEEEauthorblockA{\IEEEauthorrefmark{6}tlin@idt.eitech.edu.cn, lhe@ee.ucla.edu}
}

\IEEEoverridecommandlockouts
\IEEEpubid{\makebox[\columnwidth]{ 979-8-3503-7608-1/24\$31.00 \copyright2024 IEEE \hfill} \hspace{\columnsep}\makebox[\columnwidth]{ }}

\maketitle



\begin{abstract}

Current multimodal large language models (MLLMs) struggle to understand circuit schematics due to their limited recognition capabilities. This could be attributed to the lack of high-quality schematic-netlist training data. Existing work such as AMSnet applies schematic parsing to generate netlists. However, these methods rely on hard-coded heuristics and are difficult to apply to complex or noisy schematics in this paper. We therefore propose a novel net detection mechanism based on segmentation with high robustness. The proposed method also recovers positional information, allowing digital reconstruction of schematics. We then expand AMSnet dataset with schematic images from various sources and create AMSnet 2.0. AMSnet 2.0 contains 2,686 circuits with schematic images, Spectre-formatted netlists, OpenAccess digital schematics, and positional information for circuit components and nets, whereas AMSnet only includes 792 circuits with SPICE netlists but no digital schematics.






\end{abstract}


\section{Introduction}

Researchers employ multimodal large language models (MLLMs) in various applications in analog and mixed-signal (AMS) circuit design, such as topology design~\cite{chang2024lamagic, lai2024analogcoder, gao2025analoggenie, shi2024amsnetkg}, sizing~\cite{yin2024ado}, layout generation~\cite{liu2025layoutcopilot}, design rule check (DRC) code generation~\cite{chang2024drc}, and so on. This demonstrates that these MLLMs possess abundant field knowledge. However, current MLLMs still face difficulties in schematic recognition, understanding~\cite{bhandari2024auto}, and netlist generation. Generated data suffer from incorrectness due to hallucination, and generally cannot be used for their intended purposes, such as simulation.

\begin{figure}[t]
    \centering
    \includegraphics[width=\columnwidth]{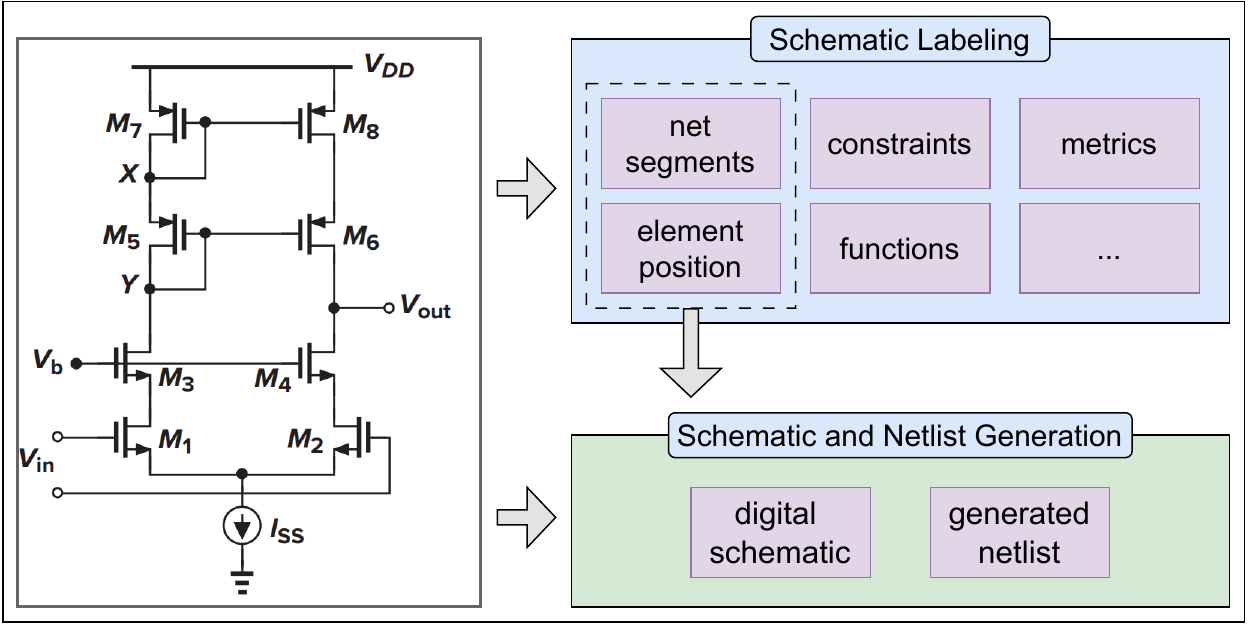}
    \caption{AMSnet 2.0 workflow for multimodal dataset construction.}
    \label{fig:pipeline}
\end{figure}

One of the main reasons for the limited ability of MLLMs to recognize circuit schematics is the lack of high-quality multi-modal training data, such as schematic-netlist pairs. Existing datasets either only focus on a single modality, such as image or language, or are too small to use for (M)LLM training. To address this issue and collect more high-quality data, we developed and released a data labeling platform, where the user uploads schematics and labels their schematic elements, wires, and expert insights, as shown in Fig.~\ref{fig:pipeline}. Backed by this platform, we construct AMSnet 2.0, a large-scale AMS dataset containing schematics, netlists, and position information for schematic elements and nets. 


\begin{figure}[t]
    \centering
    \includegraphics[width=\columnwidth]{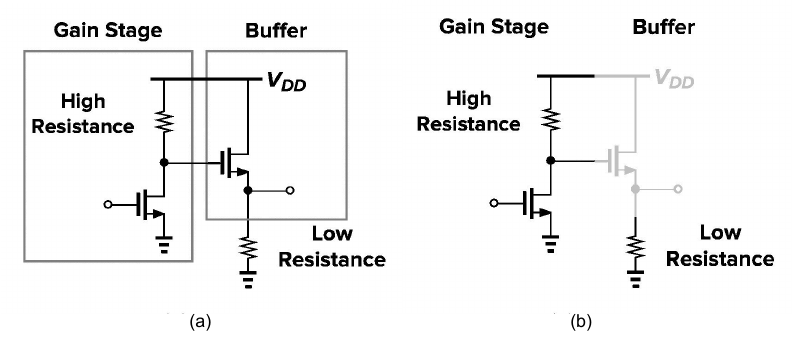}
    \caption{Examples of noisy schematics: (a) overlaid markings and (b) partial highlighting}
    \label{fig:edge_cases}
\end{figure}

In addition to manual annotations of circuit information, establishing the correspondence between schematic images and netlists is crucial for the dataset. However, the process often relies on manually interpreting images and then writing the netlists, which was time-consuming and error-prone. To address this issue, previous work~\cite{tao2024amsnet, bhandari2024auto} proposes image processing-based methods to generate netlists from schematics to naturally form desired multi-modal pairs. Some methods include template matching~\cite{lewis1995fast} to detect schematic elements and the Hough transform~\cite{illingworth1988survey} to identify wires. These methods are able to establish connectivity between schematic elements and generate netlists. However, they generally require schematic images to be clear and free of additional markings, which presents deficiencies in efficiency and robustness. It is difficult for these methods to process \textit{noisy} images, as shown in Fig.~\ref{fig:edge_cases}.

On the other hand, the above methods process wire pixels into nets via graph search algorithms or image transformations. The problem with these methods is that they only observe wire pixels as is and do not use any \textit{contextual information}, such as where the elements are and what they connect to. As a result, they may have trouble distinguishing wires from other illustrative markings (e.g. the boxes in Fig.~\ref{fig:edge_cases} (a) can be considered as wires). To compensate, these studies may assume that markings tend to show up in different colors, and then use techniques such as binarization to filter them out in a preprocessing step. However, this could introduce additional issues with a different type of illustration, as shown in Fig.~\ref{fig:edge_cases} (b), where parts of the schematic may be incorrectly filtered.


This paper presents a more robust method for net detection. It uses a deep-learning-based image instance segmentation model to accurately determine whether lines represent wires based on contextual information. By offloading the decision process to the training dataset, our method effectively differentiates wires from markings and is significantly more robust than hard-coded heuristics. We evaluate it on the AMSnet 2.0 testing set, which we divide into three difficulty levels based on schematic complexity, presence of illustrative markings, and image quality. Using a confusion matrix to analyze element types and connections in the netlist, we achieve F1 scores of 90.19\%, 84.17\%, and 80.39\% for netlist generation across these splits, respectively.

Furthermore, instance segmentation provides us with precise positional information about wires, allowing us to perform skeletonization to extract key coordinates such as ends, diverges, and turns, and summarize nets as sets of line segments. This enables the automatic reconstruction of digital schematics in widely-used formats such as OpenAccess, so that engineers can conveniently access schematics in EDA software, without having to manually modify netlists.

To summarize, our contributions in this work are as follows.
\begin{itemize}
    \item We construct and release AMSnet 2.0, a large-scale AMS dataset containing image and digital schematics, netlist, and position information for schematic elements and wires. 
    \item We develop a more robust algorithm for schematic parsing and netlist generation, eliminating the need for hard-coded heuristics in net detection.
    \item We develop a labeling platform for AMS schematics, and release it at \href{http://111.229.103.50/welcome}{\color{blue}{this link}}. It can be used to expand AMSnet 2.0.
\end{itemize}

The remainder of this paper is organized as follows. Section II introduces the schematic labeling platform. Section III presents the algorithms for schematic and netlist generation based on circuit images. Section IV shows the experimental results. Finally, Section V discusses potential future work based on AMSnet 2.0 and concludes.

\section{Schematic Labeling}
The section describes the construction process for AMSnet 2.0, which includes image collection, schematic element and net detection, and netlist generation. We collect 2686 circuit schematics from textbooks~\cite{razavi2005design} and public competitions~\cite{eda2024}, and manually annotate them using our platform. Fig.~\ref{fig:pipeline} presents an overview.

\paragraph{Element Labeling}
Users upload circuit schematics and draw bounding boxes for schematic elements on the graphical interface. They then assign the correct category labels (e.g., PMOS). The dataset is in YOLO format, including the coordinates of the object center and its width and height. Following~\cite{eda2024}, we also annotate the cross-points of all wires, categorizing whether they are visually intersecting or functionally connected.




\paragraph{Net Labeling}
Users label all the nets in a schematic image. Specifically, a net consists of one or more line segments. The users label each line segment by tracing them on the graphical interface, using a different net instance for each net. This groups all segments of the same net under the same instance, distinguishing them from other nets. The dataset contains each net and the segments it comprises, including the coordinates (start and end points) of each segment.

\paragraph{Insight Labeling}
Engineers and experts can label the type (e.g. OPAMP), name (e.g. cascode), function (e.g. signal amplification), and characteristics (e.g. high gain) of the circuit topology based on the schematic. We store this data in the form of key-value pairs. These expert insights will play a crucial role in driving the MLLM-driven automatic AMS design in the future.

\begin{figure}[t]
    \centering
    \includegraphics[width=0.9\columnwidth]{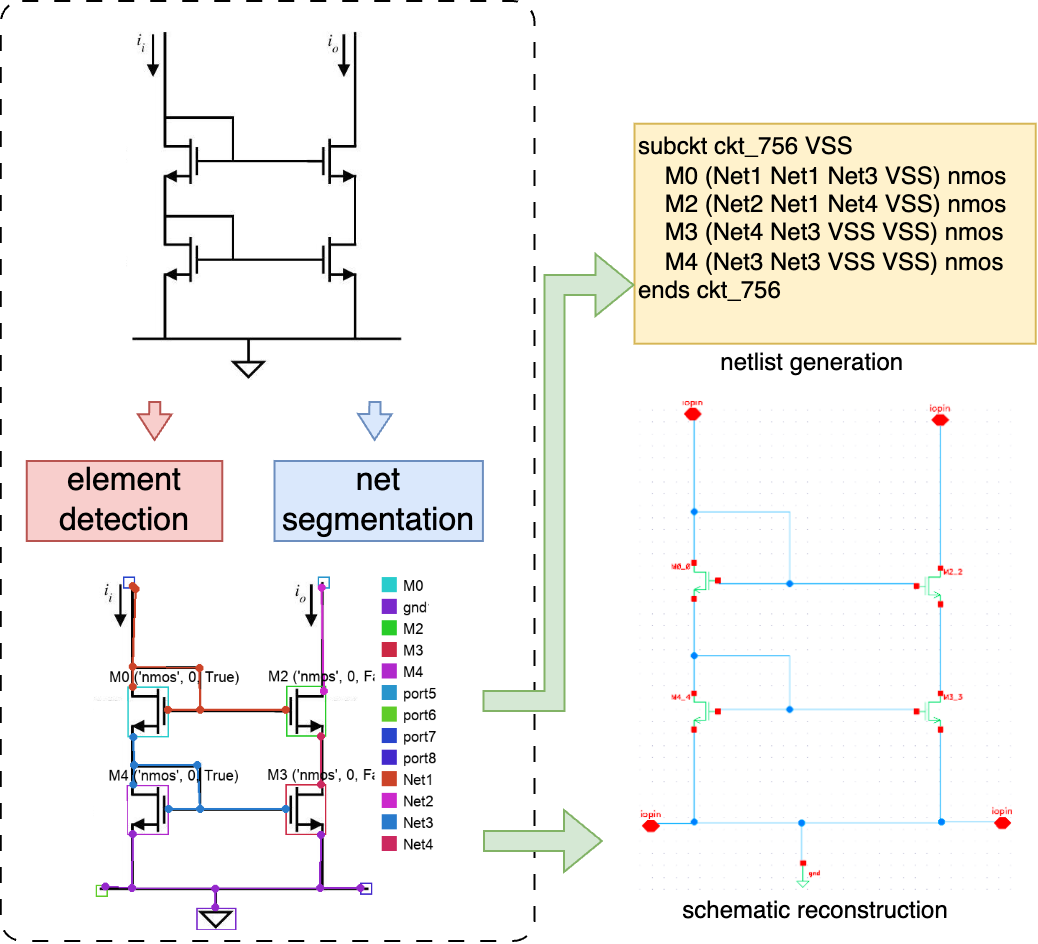}
    \caption{The full schematic processing pipeline: schematic element detection, net segmentation, Spectre format netlist generation, and digital schematic reconstruction in OpenAcess format.}
    \label{fig:method}
\end{figure}

\section{Schematic and Netlist Generation}
Fig.~\ref{fig:method} presents the proposed methods of netlist and schematic generation. We use a deep-learning-based vision model to recognize the \textit{schematic elements (i.e. voltage sources, capacitors, transistors, amplifier symbols, ground symbols, etc.)} and nets in the input images. Based on the recognition results, our method automatically reconstructs the digital schematics and generates the corresponding netlists.

\begin{figure*}[h]
    \centering
    \includegraphics[width=0.9\textwidth]{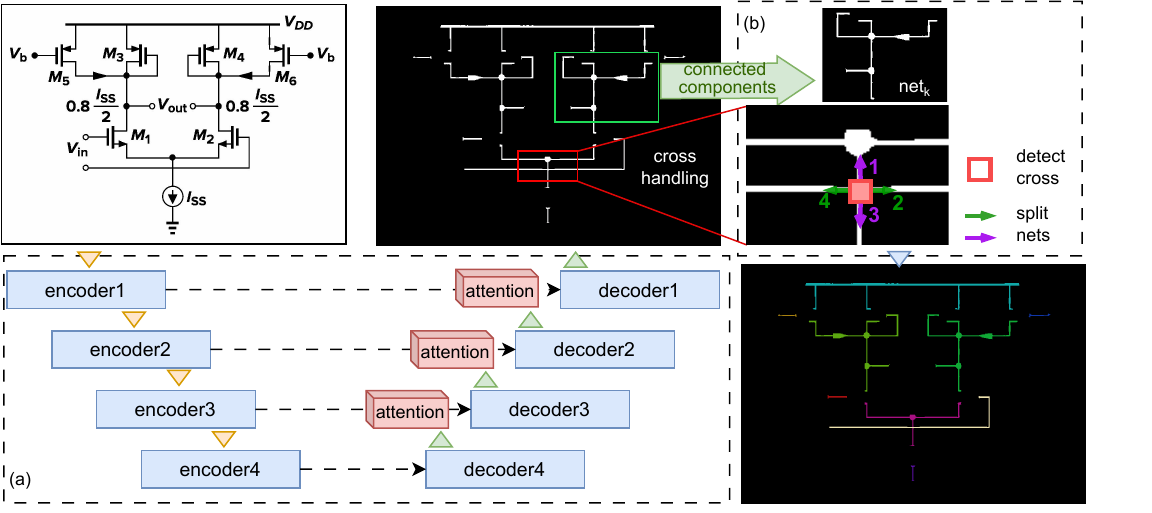}
    \caption{The proposed two-stage net detection method. (a) Semantic segmentation on the circuit wires to delineate the wires and background areas. (b) Detecting connected components, then perform split and merge operations at intersection points to identify the nets.}
    \label{fig:unet}
\end{figure*}

\subsection{Device Detection}
As in previous work~\cite{tao2024amsnet, eda2024, bhandari2024auto, shi2024amsnetkg} demonstrates, SOTA object detection models are able to effectively identify schematic elements. However, our approach extends beyond previous efforts by also detecting and recognizing the intersections of circuit nets. Specifically, when two wires meet at a cross-point, the presence of a junction (i.e. a dot) indicates that the wires belong to the same net; if no junction is present, we consider them separate nets.

\subsection{Wire Segmentation}
Previous work on wire segmentation and net detection often relies on \textit{proposal}-based instance segmentation methods~\cite{bolya2019yolact, yolo11_ultralytics}. These methods first generate potential bounding box proposals and then apply segmentation through a dedicated module. However, they struggle to fully encapsulate all pixels of a wire within a single proposal and may include pixels from multiple nets, leading to suboptimal segmentation. Additionally, \cite{hafiz2020survey} explores end-to-end instance segmentation techniques, but it still faces challenges in accurately distinguishing intersecting nets.

To address these limitations, we propose a more robust and efficient two-stage approach. The flow starts with a semantic segmentation network (U-net~\cite{ronneberger2015unet}) to segment all wires and produce \textit{wire masks}, as shown in Fig.~\ref{fig:unet}(a). The extracted wire masks are then post-processed to accurately separate intersecting nets, as shown in Fig.~\ref{fig:unet}(b). Our method eliminates the need for bounding box proposals, substantially improving segmentation accuracy and ensuring precise net delineation.

\subsection{Mocked Marking Data Augmentation}
Though the aforementioned methods perform well on \textit{clean} schematics, they still struggle in net segmentation for \textit{noisy} schematics. An example is shown in Fig.~\ref{fig:edge_cases} (a), where rectangles are drawn over the schematics. These \textit{noises} are usually overlay markings for illustration purposes; however, they significantly interfere with schematic interpretation.

Data augmentation is a standard practice in machine learning, where raw training data is altered to improve the robustness of the trained model. In this case, given sufficient training data and a reasonable data distribution, the proposed model is capable of handling data noise natively. We therefore augment raw schematics with rectangles and text markings, without changing the net mask labeling. We randomly generate \textit{mocked} markings over the schematics, and ensure that the placements of the rectangles do not directly cross over the schematic elements. After this augmentation step, the trained model is able to successfully detect wires through the mocked markings and handle these special cases.

\begin{figure*}[h]
    \centering
    \includegraphics[width=0.8\textwidth]{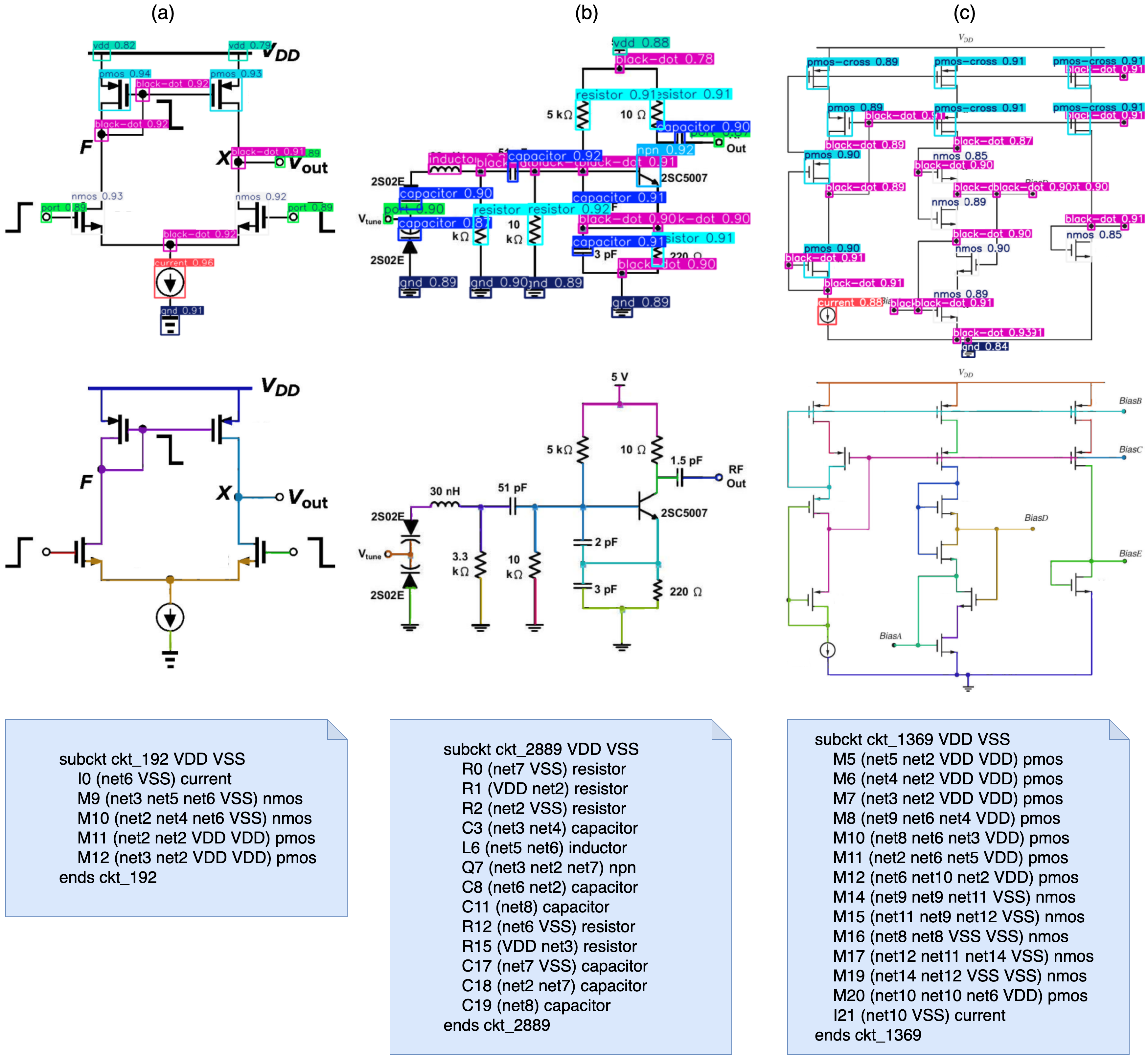}
    \caption{Results for element detection (top), net detection (middle), and Spectre format netlist generation (bottom), from the UNet procedure. Columns (a), (b), and (c) present examples from easy, medium, and hard splits respectively.}
    \label{fig:schematic_difficulties}
\end{figure*}

\subsection{Net Recognition}
 After device and wire recognition, the flow needs to understand how they are connected to form nets. Given the segmented wire masks from the above steps, those that do not intersect each other belong to separate connected components. We assign a unique net label to each connected component. Fig.~\ref{fig:unet} (b) shows an example of wire masks that interconnect and are affected by cross lines. In this case, a mask is applied to the area around the intersection points, which splits the connected component at the intersection. Then, the tool merges the opposite connections to precisely identify the locations of multiple nets. Specifically, it orders all four connections to the intersection based on the angle between the connecting points and the center of the intersection, as shown in the green and purple numbers {1, 2, 3, 4} in Fig.~\ref{fig:unet}(b). Finally, the connections {1, 3} and {2, 4} are grouped to resolve the intersection.
    

\subsection{Netlist Generation \& Schematic Reconstruction}
After detecting each schematic element and its rotation angle, we can identify the pin areas for each element (i.e., the pins of a resistor are located at the midpoint of each end of its bounding box). By identifying the nearest or intersecting net labels for each element and its corresponding pins, the tool can establish the connectivity relationship and generate a precise netlist. 

Up to this stage, the tool has collected the coordinates of the schematic elements and net segments to reconstruct editable digital schematics in widely-used formats, such as OpenAccess. This process ensures that the generated electronic schematics are readable without the need for manual circuit construction and allows for further modifications and simulations.


\begin{figure*}[h]
    \centering
    \includegraphics[width=0.9\textwidth]{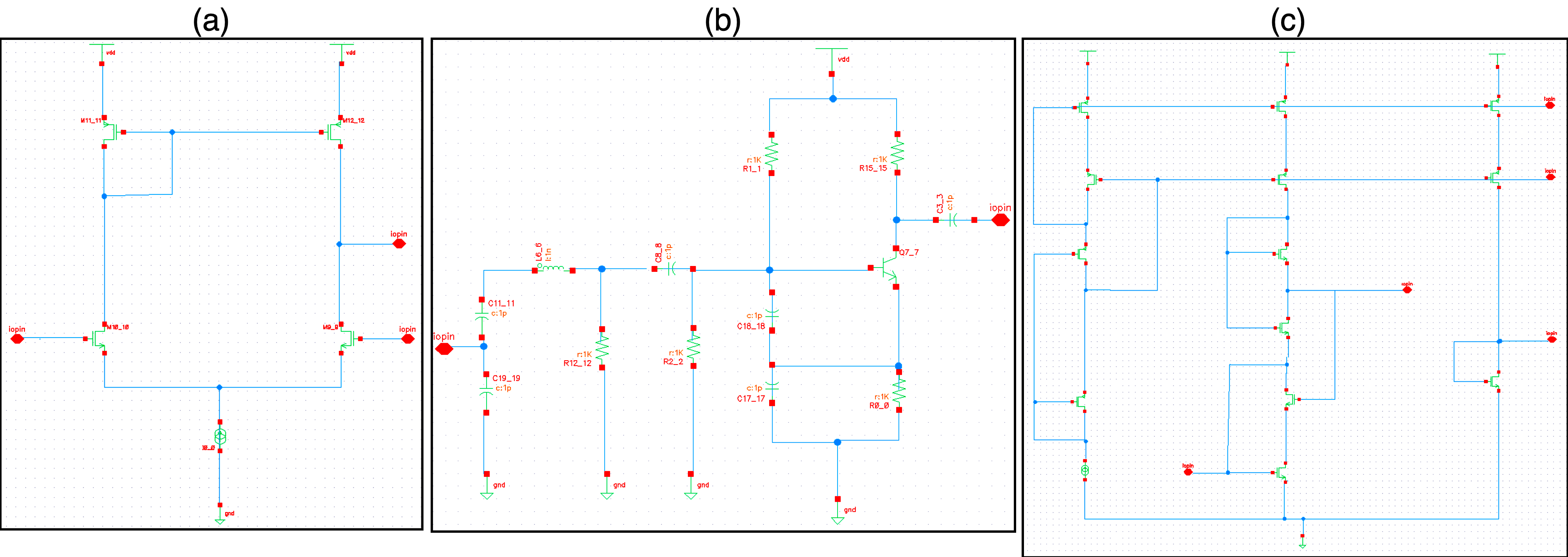}
    \caption{Generated schematics in OpenAccess format}
    \label{fig:oa}
\end{figure*}

\section{Experiments}
\label{sec:experiments}
\subsection{Experiment Setting}
We train the models on labeled data with a training – validation - testing split of 1986 - 500 - 200 schematics. We employ YOLO11 for element detection, training it for 500 epochs with a batch size of 16, along with other default parameters of YOLO11. For wire segmentation, we use two models, UNet and YOLO11-seg, and compare their results. We train UNet for 1000 epochs with a batch size of 32, and YOLO11-seg for 1500 epochs with a batch size of 16, both with default parameters. We conduct all experiments on an NVIDIA RTX 4090 GPU.

\subsection{Evaluation Protocol}
For the evaluation of netlist generation, we select three test sets with varying levels of difficulty—easy (80), medium (70), and hard (50)—based on the complexity of the circuit topology as well as the quality and style of the images. Specifically, we manually evaluate the test set schematics on four metrics: high element count ($\ge$ 10), crossing wires, overlaid markings, and low resolution. Images satisfying none of the above are labeled ``easy'', images satisfying exactly one of the above are ``medium'', and the rest are ``hard''. Fig.~\ref{fig:schematic_difficulties} presents three examples, each from the easy, medium, and hard splits, along with resulted element detection, net detection, and Spectre netlist generated by the UNet-based procedure.

\subsection{Netlist Generation Results}
As previously mentioned, we apply two options for the wire detection in our experiments. One is YOLO11-seg~\cite{yolo11_ultralytics}, which directly detects wires into separate nets. The other is UNet, which first detects wire chunks, and then algorithmically post-processes them into separate nets with the help of crossing points detected during element detection. We therefore report two sets of results, as shown in Table~\ref{tab:main_results}.

The reported scores are evaluated via a confusion matrix. For each netlist component (schematic elements excluding VDD, GND, ports, etc.) and net connection \textit{in the ground truth netlist}, if it was correctly predicted in the predicted netlist, we mark this as a true positive. If it was missing or predicted incorrectly, we mark this as a false negative. Finally, if the element or connection does not exist but was predicted (i.e. extra capacitor, or extra connection on a component, etc.), we mark this as a false positive. From this point, we use the standard precision-recall-F1 definition, and report the F1 scores in Table~\ref{tab:main_results}. Note that incorrect type prediction, such as predicting capacitors as resistors, is also considered a false negative.


\begin{table}[t]
\centering
\resizebox{0.8\columnwidth}{!}{ 
\begin{tabular}{lccc}
\hline
Model & Easy & Medium & Hard  \\ \hline
Unet~\cite{ronneberger2015unet} & 90.19 & 84.17 & 80.39  \\
YOLO11-seg~\cite{yolo11_ultralytics} & 83.62 & 71.10 & 73.11  \\ \hline
\end{tabular}
}
\caption{Netlist generation F1 scores}
\label{tab:main_results}
\end{table}


\vspace{0.5mm}
Since different netlists could have different element names and net names, direct comparison may undermine the results; we therefore consider all permutations between ground truth names and predicted names. For example, for netlist ckt\_192 shown in Fig.~\ref{fig:schematic_difficulties} (a), the element names for the predicted netlist include I0 and M9-12, and the net names include VDD, VSS, and net2-6. Suppose the ground truth VDD is labeled ``net1'', the correct permutation should match net1 against VDD among others. We report the F1 score generated by the best possible permutation. Since the computation complexity for permutations grows exponentially, we manually determine the F1 score when the schematics are too complex.




\subsection{Schematic Reconstruction Results}
Fig.~\ref{fig:oa} presents the schematic reconstruction results for the images shown in Fig.~\ref{fig:schematic_difficulties}. The automatic generation files are in Open Access database format and displayed in Cadence Virtuoso. Compared with the original images, we can see that our flow has successfully reconstructed the digital schematics for AMSnet 2.0. These automatically reconstructed schematics accurately reproduce the element layout, net positions, and connections depicted in the images, requiring minimal manual intervention.

\begin{figure*}[h]
    \centering
    \includegraphics[width=0.9\textwidth]{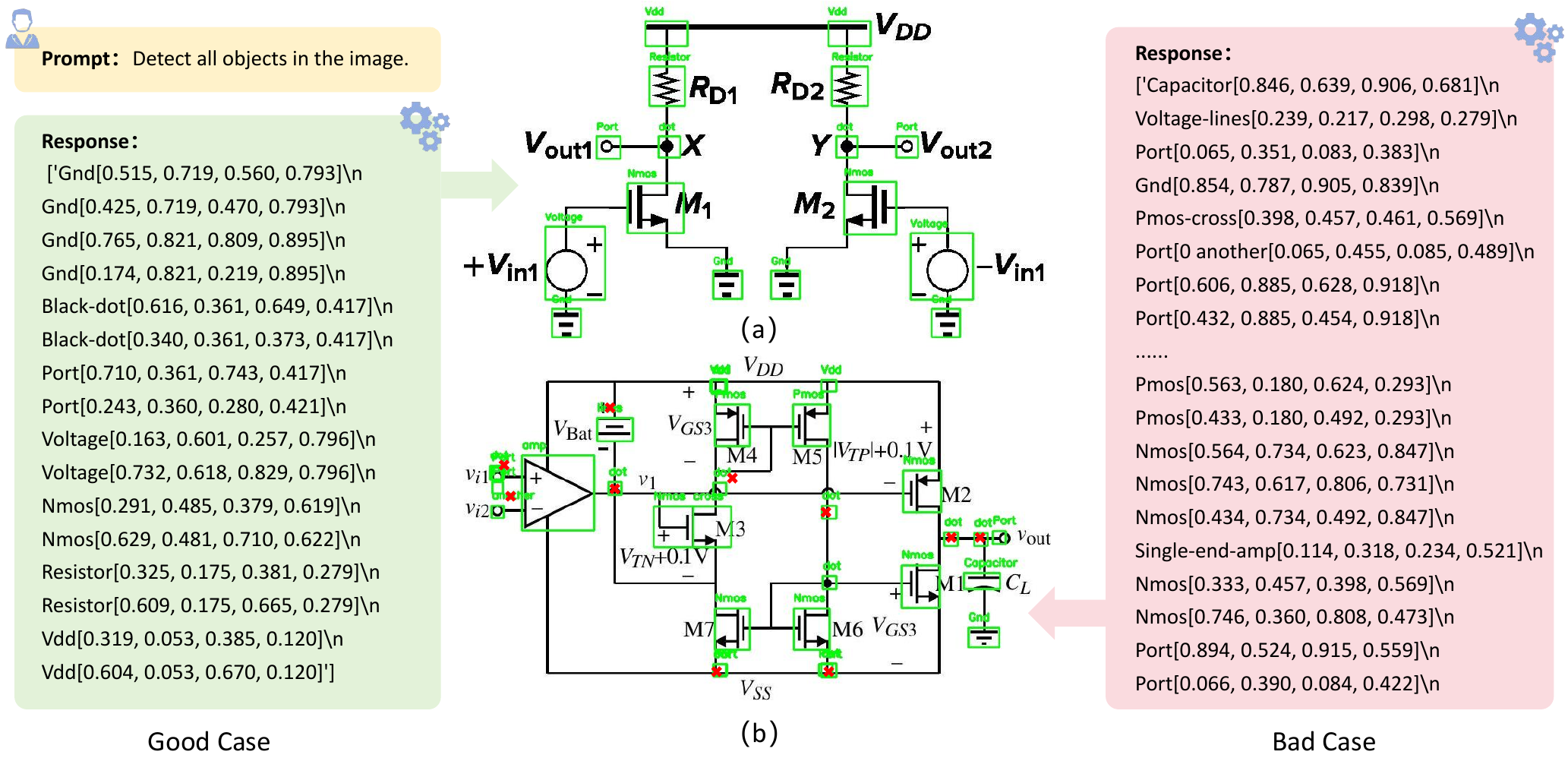}
    \caption{LLM-based schematic analysis after SFT based on AMSnet 2.0}
    \label{fig:det}
\end{figure*}

\section{Conclusions and Discussions}
In this work, we introduce AMSnet 2.0, a larger-scale dataset that includes schematics images, Spectre format netlists, and position information. We propose a method based on instance segmentation for net detection. Utilizing the detected elements and net position information, we implement an automatic tool that can generate netlists and  reconstruct schematics in OpenAccess format. We release our data annotation platform, allowing the community to annotate circuit data and further enlarge the dataset.

\subsection{AMSnet 2.0 Statistics}
\begin{figure}[t]
    \centering
    \includegraphics[width=0.42\textwidth]{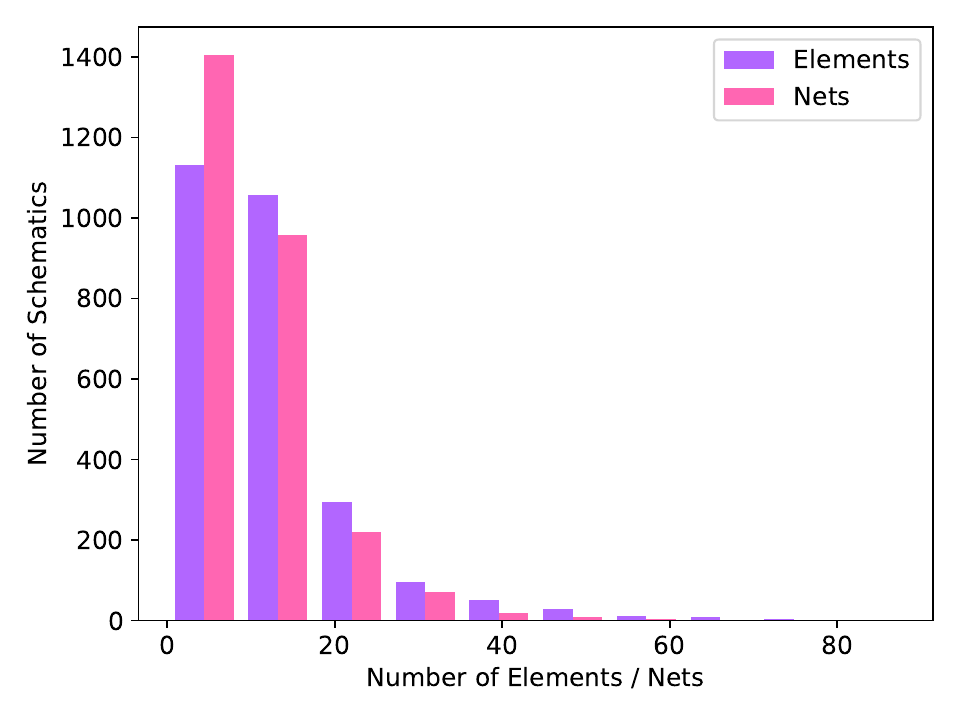}
    \vspace{-2.5mm}
    \caption{Distribution of schematic complexity by numbers of elements and mets}
    \label{fig:num}
    \vspace{-5.5mm}
\end{figure}

\begin{figure}[t]
    \centering
    \includegraphics[width=0.42\textwidth]{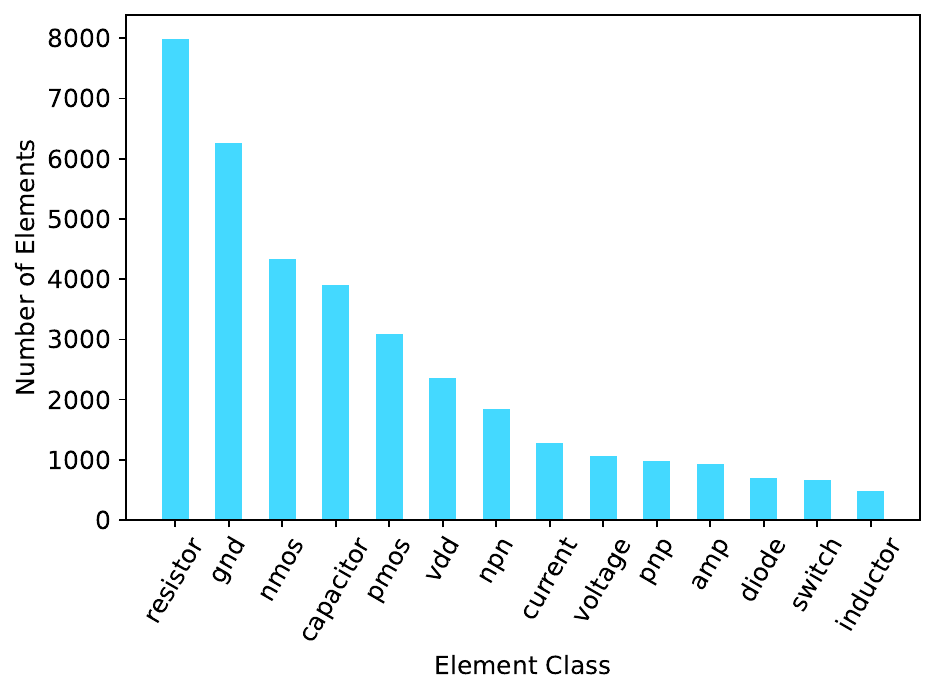}
    \vspace{-1.5mm}
    \caption{Distribution of schematic element types}
    \label{fig:class}
    \vspace{-6.5mm}
\end{figure}

Fig.~\ref{fig:num} presents the data distribution for the 2686 circuits in AMSnet 2.0. We can see that the majority of circuits include around 10-20 elements and nets, but the larger ones could contain up to 80. Fig.~\ref{fig:class} presents the element type distribution; we can see that AMSnet 2.0 contains hundreds to thousands of each type. The full dataset will be open-sourced upon the acceptance of this paper, along with corresponding Spectre format netlists, OpenAccess format schematics, and position information for all elements and net segments.

\vspace{-0.5mm}



\subsection{Future Applications of AMSnet2.0}

The multimodal AMSnet 2.0 establishes the correspondence between schematics and netlists and collects a large amount of human design experience through the annotation platform. With AMSnet 2.0 as the dataset, it is possible to empower MLLMs or other AI models to address key challenges in AMS circuit design in the future, which we will name a few in the following paragraphs.

\paragraph{Schematics Generation from Netlist}
Converting a netlist into a readable schematic can help designers quickly understand an AMS circuit, which has been a long-standing EDA challenge. With the extensive amount of schematic-netlist pairs in AMSnet 2.0, we plan to train an LLM to facilitate the conversion of netlists to readable schematic diagrams. This work will greatly ease the understanding and debugging of AMS circuits generated by LLMs. It will also enable the creation of more schematic-netlist pairs and further enlarge AMSnet 2.0.

\paragraph{LLM Enhanced for AMS Circuit Understanding}
Due to the current lack of datasets that include large-scale, high-quality AMS circuit schematics and netlists, the performance of SOTA LLMs and MLLMs in understanding circuit schematics remains suboptimal~\cite{bhandari2024auto}. Specifically, they struggle to accurately generate the circuit netlists.
Building on AMSnet 1.0, the 2.0 dataset includes a richer set of visual information with the introduction of key-point coordinates. This data is crucial for training AMS-specific MLLMs to enhance their understanding of circuit schematics. We conduct supervised fine-tuning (SFT) of the MLLM using the existing data. As shown in Fig.~\ref{fig:det}, it is evident that after SFT, the MLLM possesses basic capabilities in detecting and locating elements in simple circuit schematics, although their performance on more complex diagrams still needs improvement. 

In the future, we expect the continuous growth of AMSnet 2.0 enabled by the image-to-netlist and netlist-to-schematic pipelines. The dataset will be used for SFT and reinforcement learning in SOTA MLLMs. Our goal is to facilitate MLLMs with robust capabilities in understanding AMS circuit schematics.

\newpage





\balance
\bibliographystyle{ieeetr}
\bibliography{refs}
\balance

\end{document}